\pdfoutput=1
\documentclass[11pt]{article}
\usepackage{EMNLP2023}
\usepackage{times}
\usepackage{latexsym}
\usepackage[T1]{fontenc}
\usepackage[utf8]{inputenc}
\usepackage{microtype}
\usepackage{inconsolata}
\usepackage{lipsum}
\usepackage{graphicx}
\usepackage{tablefootnote}
\usepackage{refcount}

\newcommand\blankfootnote[1]{%
  \let\thefootnote\relax\footnotetext{#1}%
  \let\thefootnote\svthefootnote%
}

\newcommand{\tablefootnotemark}[1]%
{\textsuperscript{\getrefnumber{#1}}}

\title{Kandinsky: an Improved Text-to-Image Synthesis with \\ Image Prior and Latent Diffusion} 

\author{
Anton Razzhigaev\textsuperscript{1,2},  
Arseniy Shakhmatov\textsuperscript{3}, 
Anastasia Maltseva\textsuperscript{3},
Vladimir Arkhipkin\textsuperscript{3}, \\
\bf Igor Pavlov\textsuperscript{3},
Ilya Ryabov\textsuperscript{3}, 
Angelina Kuts \textsuperscript{3}, \bf Alexander Panchenko\textsuperscript{2,1}, \\
\bf Andrey Kuznetsov\textsuperscript{3,1}, and 
Denis Dimitrov\textsuperscript{3,1}\\
\textsuperscript{1}AIRI, 
\textsuperscript{2}Skoltech,
\textsuperscript{3}Sber AI\\
\href{mailto:razzhigaev@airi.net}{\{razzhigaev, kuznetsov, dimitrov\}@airi.net} \\ 
}

\begin{document}

\maketitle

\begin{abstract}
Text-to-image generation is a significant domain in modern computer vision and has achieved substantial improvements through the evolution of generative architectures. Among these, there are diffusion-based models that have demonstrated essential quality enhancements. These models are generally split into two categories: pixel-level and latent-level approaches. We present Kandinsky\footnote{The system is named after \href{https://en.wikipedia.org/wiki/Wassily_Kandinsky}{Wassily Kandinsky}, a famous painter and an art theorist.}, a novel exploration of latent diffusion architecture, combining the principles of the image prior models with latent diffusion techniques. The image prior model is trained separately to map text embeddings to image embeddings of CLIP. Another distinct feature of the proposed model is the modified MoVQ implementation, which serves as the image autoencoder component. Overall, the designed model contains 3.3B parameters. We also deployed a user-friendly demo system that supports diverse generative modes such as text-to-image generation, image fusion, text and image fusion, image variations generation, and text-guided inpainting/outpainting. Additionally, we released the source code and checkpoints for the Kandinsky models. Experimental evaluations demonstrate a FID score of 8.03 on the COCO-30K dataset, marking our model as the top open-source performer in terms of measurable image generation quality.
\end{abstract}

\section{Introduction}
\renewcommand{\arraystretch}{1.2}

\begin{figure*}
  \centering
  \includegraphics[width=2.1\columnwidth]{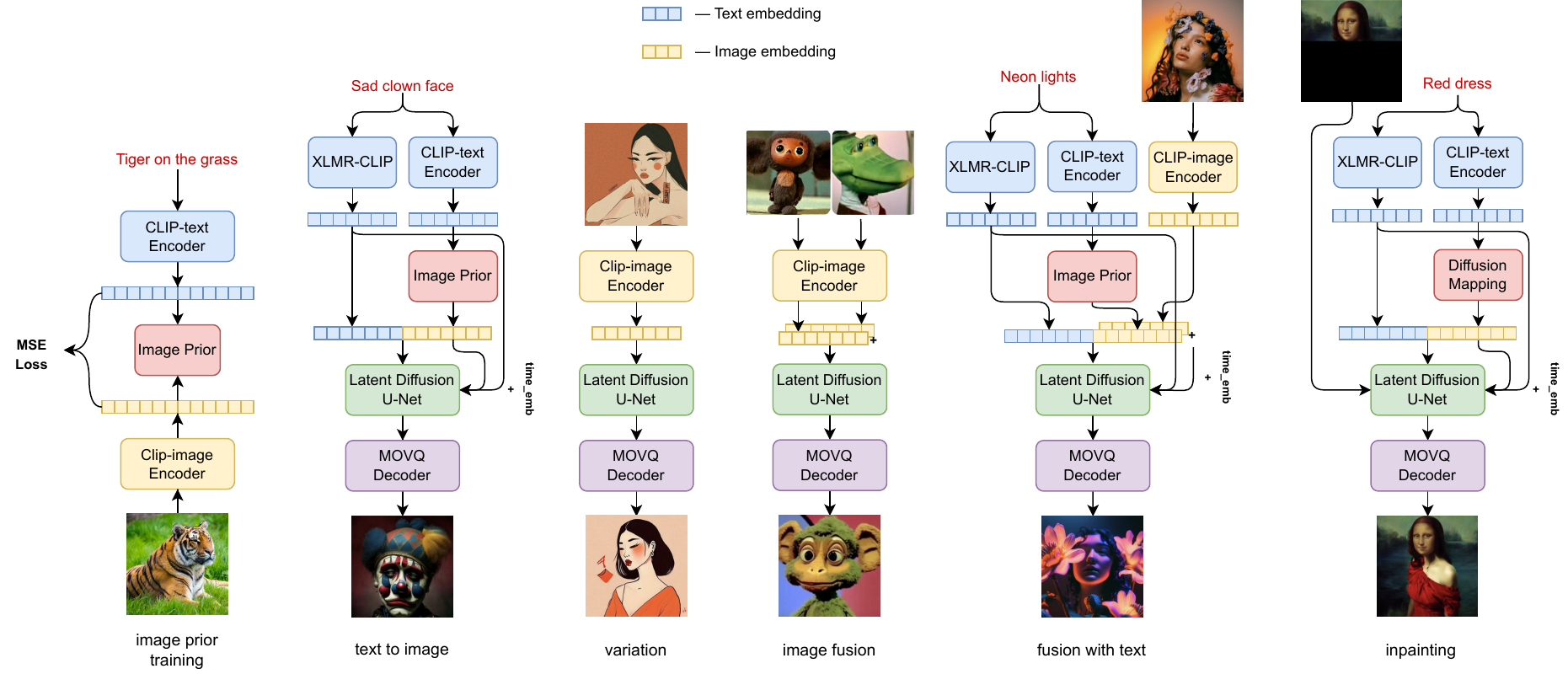}
  \caption{Image prior scheme and inference regimes of the Kandinsky model.}
  \label{fig_architecture_inference_regimes}
\end{figure*}


 In quite a short period of time, generative abilities of text-to-image models have improved substantially, providing users with photorealistic quality, near real-time inference speed, a great number of applications and features, including simple easy-to-use web-based platforms and sophisticated AI graphics editors. 

This paper presents our unique investigation of latent diffusion architecture design, offering a fresh and innovative perspective on this dynamic field of study. First, we describe the new architecture of Kandinsky and its details. The demo system with implemented features of the model is also described. Second, we show the experiments, carried out in terms of image generation quality and come up with the highest FID score among existing open-source models. Additionally, we present the rigorous ablation study of prior setups that we conducted, enabling us to carefully analyze and evaluate various configurations to arrive at the most effective and refined model design.

Our \textbf{contributions} are as follows:
  \begin{itemize}

    \item We present the first text-to-image architecture designed using a combination of image prior and latent diffusion.
    
    \item We demonstrate experimental results comparable to the state-of-the-art (SotA) models such as Stable Diffusion, IF, and DALL-E 2, in terms of FID metric and achieve the SotA score among all existing open source models.
  
    \item We provide a software implementation of the proposed state-of-the-art method for text-to-image generation, and release pre-trained models, which is unique among the top-performing methods. Apache 2.0 license makes it possible to use the model for both non-commercial and commercial purposes.\footnote{\url{https://github.com/ai-forever/Kandinsky-2}} \footnote{\url{https://huggingface.co/kandinsky-community}}  

    \item We create a web image editor application that can be used for interactive generation of images by text prompts (English and Russian languages are supported) on the basis of the proposed method, and provides inpainting/outpainting functionality.\footnote{\url{https://fusionbrain.ai/en/editor}}
    The video demonstration is available on YouTube.\footnote{\url{https://www.youtube.com/watch?v=c7zHPc59cWU}} 
    
  \end{itemize}

\section{Related Work}

Early text-to-image generative models, such as DALL-E \cite{DBLP:conf/icml/RameshPGGVRCS21} and CogView \cite{DBLP:conf/nips/DingYHZZYLZSYT21}, or later Parti \cite{DBLP:journals/tmlr/YuXKLBWVKYAHHPLZBW22} employed autoregressive approaches but often suffered from significant content-level artifacts. This led to the development of a new breed of models that utilized the diffusion process to enhance image quality. Diffusion-based models, such as DALL-E 2 \cite{ramesh2022hierarchical}, Imagen \cite{saharia2022photorealistic}, and Stable Diffusion\footnote{\url{https://github.com/CompVis/stable-diffusion}}, have since become cornerstones in this domain. These models are typically divided into pixel-level \cite{ramesh2022hierarchical, saharia2022photorealistic} and latent-level \cite{rombach2022highresolution} approaches.

This surge of interest has led to the design of innovative approaches and architectures, paving the way for numerous applications based on open-source generative models, such as DreamBooth \cite{ruiz2022dreambooth} and DreamPose \cite{karras2023dreampose}. These applications exploit image generation techniques to offer remarkable features, further fueling the popularity and the rapid development of diffusion-based image generation approaches.

This enabled a wide array of applications like 3D object synthesis \cite{DBLP:conf/iclr/PooleJBM23, DBLP:journals/corr/abs-2303-14184, DBLP:journals/corr/abs-2211-10440, DBLP:journals/corr/abs-2303-13873}, video generation \cite{DBLP:conf/nips/HoSGC0F22, DBLP:journals/corr/abs-2303-08320, DBLP:journals/corr/abs-2210-02303, DBLP:conf/iclr/SingerPH00ZHYAG23, DBLP:journals/corr/abs-2304-08818, DBLP:journals/corr/abs-2302-03011}, controllable image editing \cite{DBLP:conf/iclr/HertzMTAPC23, DBLP:conf/siggraph/ParmarS0LLZ23, DBLP:journals/corr/abs-2210-16056, DBLP:journals/corr/abs-2307-02421, DBLP:journals/corr/abs-2307-12493}, and more, which are now at the forefront of this domain.

\begin{figure*}
  \centering
  \includegraphics[width=2.1\columnwidth]{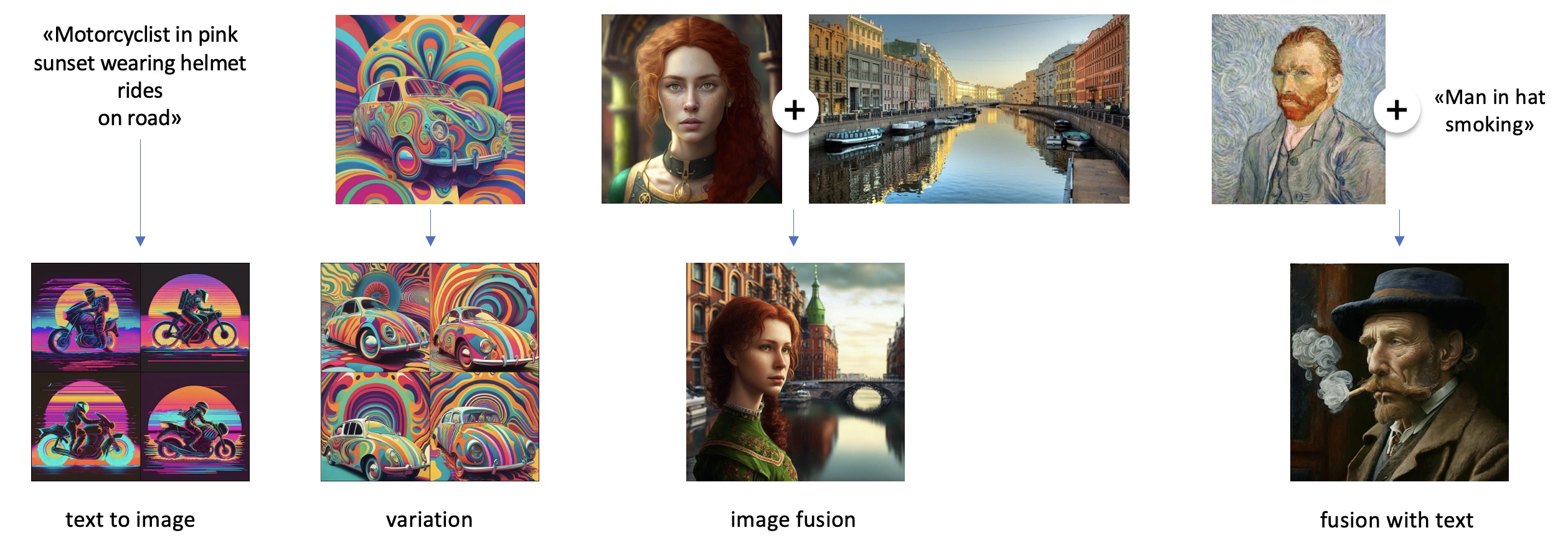}
  \caption{Examples of inference regimes using Kandinsky model.}
  \label{fig_inference_regimes}
\end{figure*}

Diffusion models achieve state-of-the-art results in image generation task both unconditional \cite{ho2020denoising, nichol2021improved} and conditional \cite{peebles2022scalable}. They beat GANs \cite{NIPS2014_5ca3e9b1} by generating images with better scores of fidelity and diversity without adversarial training \cite{dhariwal2021diffusion}. Diffusion models also show the best performance in various image processing tasks like inpainting, outpainting, and super-resolution \cite{batzolis2021conditional, saharia2022palette}.

Text-to-image diffusion models have become a popular research direction  due to the high performance of diffusion models and the ability to simply integrate text conditions with the classifier-free guidance algorithm \cite{ho2021classifierfree}. Early models like GLIDE \cite{DBLP:conf/icml/NicholDRSMMSC22}, Imagen \cite{saharia2022photorealistic}, DALL-E 2 \cite{ramesh2022hierarchical} and eDiff-I \cite{balaji2023ediffi} generate low-resolution image in pixel space and then upsample it with another super-resolution diffusion models. They are also using different text encoders, large language model T5 \cite{raffel2020exploring} in Imagen, CLIP \cite{DBLP:conf/icml/RadfordKHRGASAM21} in GLIDE and DALL-E 2.

\section {Demo System}
We implemented a set of user-oriented solutions where Kandinsky model is embedded as a core imaging service. It has been done due to a variety of inference regimes, some of which need specific front-end features to perform properly. Overall, we implemented two main inference resources: \href{https://t.me/kandinsky21_bot}{Telegram bot} and \href{https://editor.fusionbrain.ai}{FusionBrain website}.

\begin{figure*}
  \centering
\includegraphics[height=6.2cm, width=1.2\columnwidth, keepaspectratio]{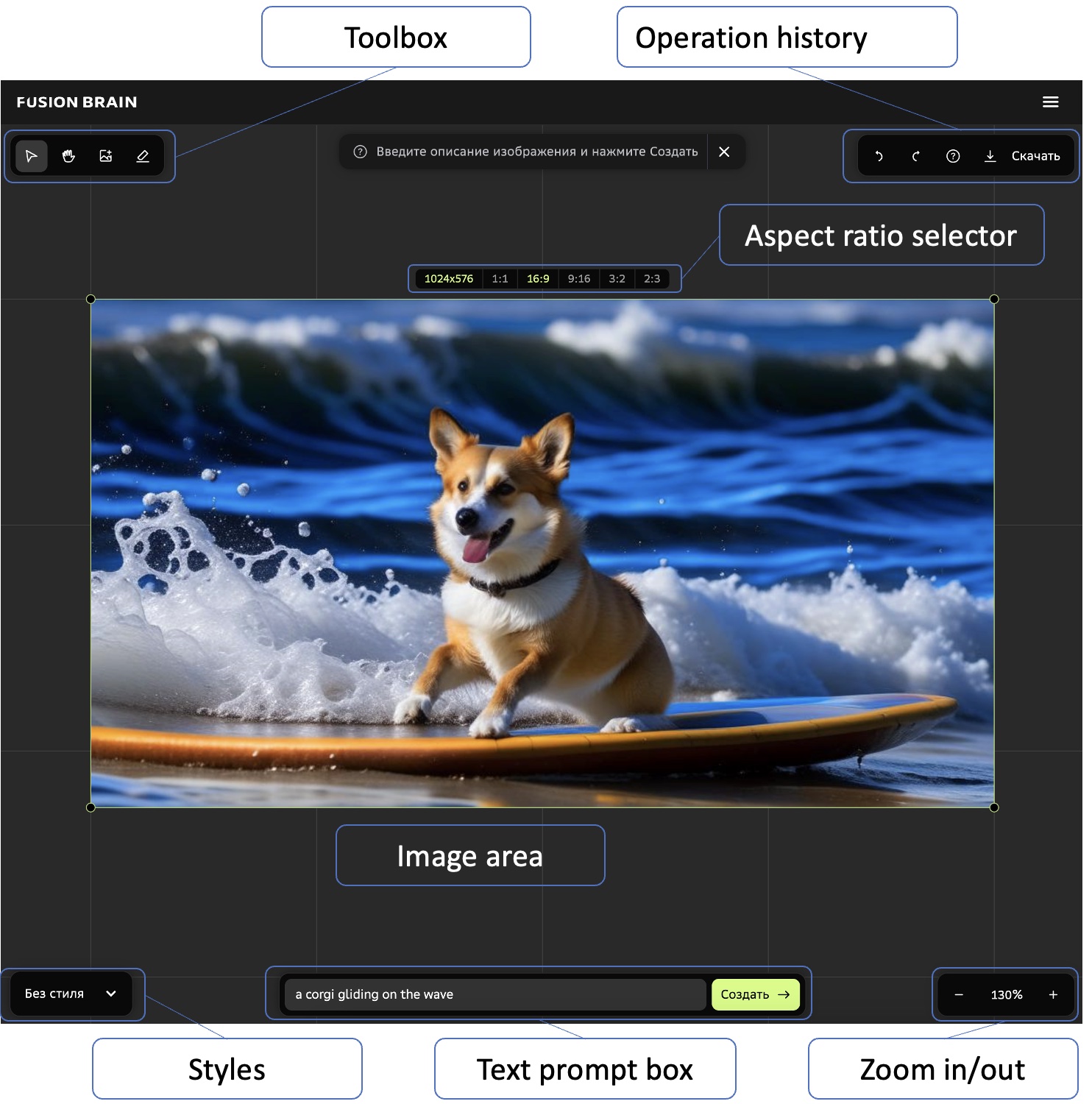}
\includegraphics[width=1.2\columnwidth]{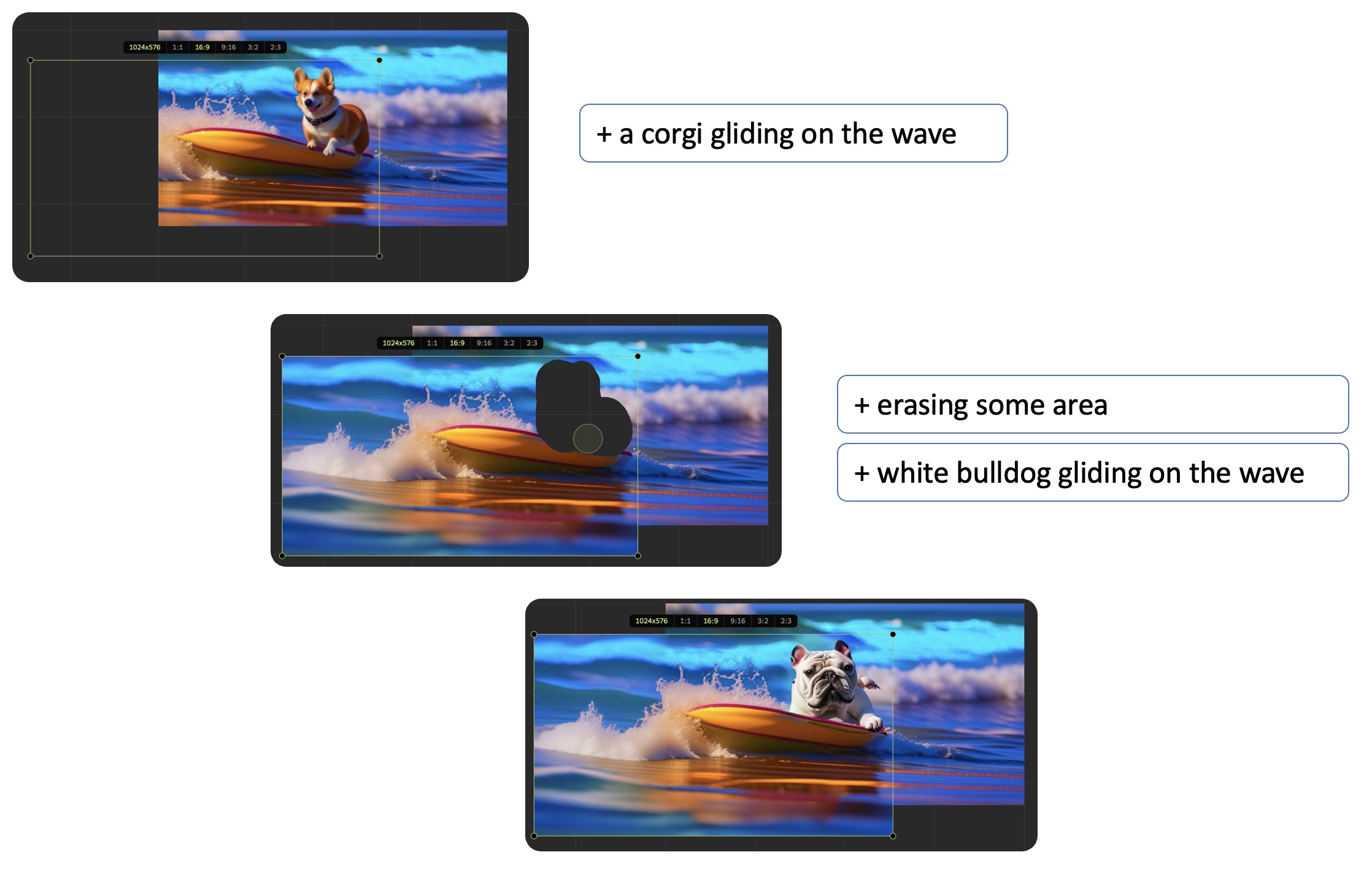}
    
  \caption{Kandinsky web interface for ``a corgi gliding on the wave'': generation (left) and in/outpainting (right).}
  \label{fusionbrain UI}
\end{figure*}

FusionBrain represents a web-based image editor with such features as loading and saving images, sliding location window, erasing tools, zooming in/out, various styles selector, etc. (cf. Figure ~\ref{fusionbrain UI}). In terms of image generation, the three following  options are implemented on this side:
\begin{itemize}
    \item text-to-image generation -- user inputs a text prompt in Russian or English, then selects an aspect-ratio from the list (9:16, 2:3, 1:1, 16:9, 3:2), and the system generates an image;
    \item inpainting -- using the specific erasing tool, user can remove any arbitrary input image part and fill it, guided by a text prompt or without any guidance;
    \item outpainting -- input image can be extended with a sliding window that can be used as a mask for the following generation (if the window intersects any imaged area, then the empty window part is generated with or without text prompt guidance).
\end{itemize}

Inpainting and outpainting options are the main image editing features of the model. Architectural details about these generation types can also be found in Figure~\ref{fig_architecture_inference_regimes}.

%

Telegram bot contains the following image generation features (cf. Figure~\ref{fig_inference_regimes}):
\begin{itemize}
    \item text-to-image generation;
    \item image and text fusion -- user inputs an image and a text prompt to create a new image guided by this prompt;
    \item image fusion -- user inputs an image as the main one and another 'guiding' image, and the system generates their fusion;
    \item image variations -- user inputs an image, and the system generates several new images similar to the input one.
\end{itemize}

\section{Kandinsky Architecture}

In our work, we opted to deliver state-of-the-art text-to-image synthesis. In the initial stages of our research, we experimented with multilingual text encoders, such as mT5 \cite{DBLP:conf/naacl/XueCRKASBR21}, XLMR \cite{DBLP:conf/acl/ConneauKGCWGGOZ20}, XLMR-CLIP\footnote{\url{https://github.com/FreddeFrallan/Multilingual-CLIP}}, to facilitate robust multilingual text-to-image generation. However, we discovered that using the CLIP-image embeddings instead of standalone text encoders resulted in improved image quality. As a result, we adopted an image prior approach, utilizing diffusion and linear mappings between text and image embedding spaces of CLIP, while keeping additional conditioning with XLMR text embeddings. That is why Kandinsky uses two text encoders: CLIP-text with image prior mapping and XLMR. We have set these encoders to be frozen during the training phase.

The significant factor that influenced our design choice was the efficiency of training latent diffusion models, as compared to pixel-level diffusion models \cite{rombach2022highresolution}. This led us to focus our efforts on the latent diffusion architecture. Our model essentially comprises three stages: text encoding, embedding mapping (image prior), and latent diffusion.

The construction of our model involves three primary steps: text encoding, embedding mapping (image prior), and latent diffusion.
At the embedding mapping step, which we also refer to as the image prior, we use the transformer-encoder model. This model was trained from scratch with a diffusion process on text and image embeddings provided by the CLIP-ViT-L14 model. A noteworthy feature in our training process is the use of element-wise normalization of visual embeddings. This normalization is based on full-dataset statistics and leads to faster convergence of the diffusion process. We implemented inverse normalization to revert to the original CLIP-image embedding space in the inference stage.

\begin{table}
\small 
\centering
  \caption{Proposed architecture comparison by FID on COCO-30K validation set on $256 \times 256$ resolution. *~For the IF model we reported reproduced results on COCO-30K, but authors provide FID of 7.19.}
  \label{fid_kandinsky}
    \begin{tabular}{lc}
    \hline
     \textbf{Model} & \textbf{FID-30K}\\
    \hline
    \multicolumn{2}{c}{\textit{Open Sourced Techologies}} \\
    \hline
        \textbf{Kandinsky (Ours)} & \textbf{8.03} \\
        Stable Diffusion 2.1 (2022) \tablefootnote{\label{SD}\url{https://github.com/Stability-AI/stablediffusion}} & 8.59 \\
        GLIDE  \textsuperscript{\getrefnumber{SD}} \cite{DBLP:conf/icml/NicholDRSMMSC22} & 12.24 \\
        IF* (2023) \textsuperscript{\getrefnumber{IF}} & 15.10 \\
        Kandinsky 1.0 (2022) \tablefootnote{\label{rudalle}\url{https://github.com/ai-forever/ru-dalle}} & 15.40 \\
        ruDALL-E Malevich (2022) \textsuperscript{\getrefnumber{rudalle}} & 20.00 \\
        GLIGEN \tablefootnote{\url{https://github.com/gligen/GLIGEN}} \cite{DBLP:journals/corr/abs-2301-07093} & 21.04 \\
        \hline
        \multicolumn{2}{c}{\textit{Proprietary Technologies}} \\
        \hline
        \textbf{eDiff-I \cite{balaji2023ediffi}} & \textbf{6.95} \\
        Imagen \cite{saharia2022photorealistic} & 7.27 \\
        GigaGAN \cite{kang2023scaling} & 9.09 \\
        DALL-E 2 \cite{ramesh2022hierarchical} & 10.39 \\
        DALL-E \cite{DBLP:conf/icml/RameshPGGVRCS21} & 17.89 \\
    \hline
    \end{tabular}
\end{table}

\begin{figure*}
  \centering
  \includegraphics[width=2.\columnwidth]{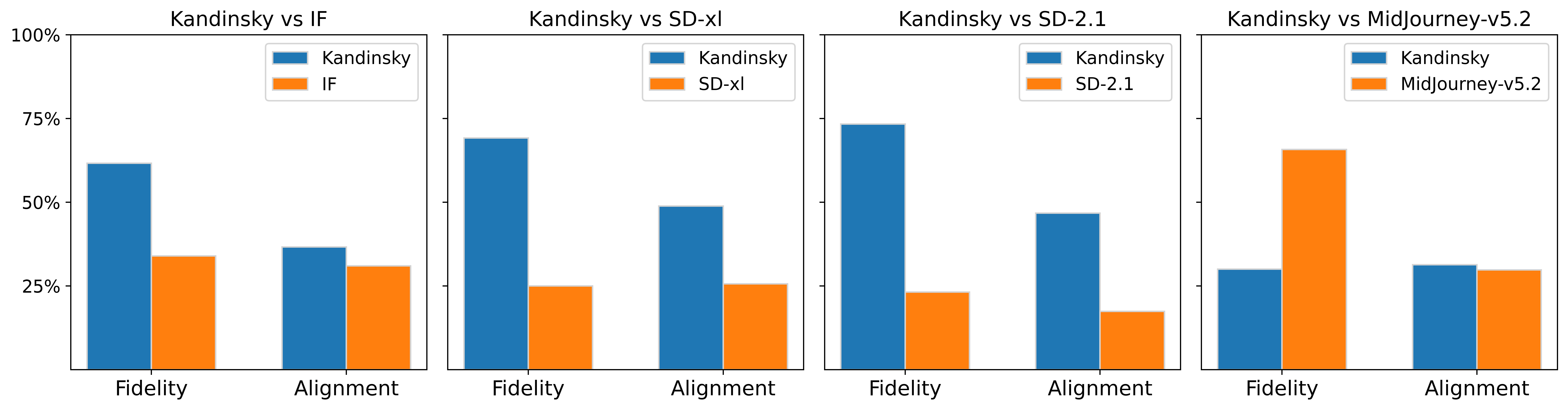}
  \caption{Human evaluation: competitors vs Kandinsky with diffusion prior on Drawbench. The total count of votes is 5000.}
  \label{human_eval}
\end{figure*}

The image prior model is trained on text and image embeddings, provided by the CLIP models. We conducted a series of experiments and ablation studies on the specific architecture design of the image prior model (Table \ref{fid_clip_table}, Figure \ref{human_eval}). The model with the best human evaluation score is based on a 1D-diffusion and standard transformer-encoder with the following parameters: num\_layers=20, num\_heads=32, and hidden\_size=2048.

The latent diffusion part employs a UNet model along with a custom pre-trained autoencoder. Our diffusion model uses a combination of multiple condition signals: CLIP-image embeddings, CLIP-text embeddings, and XLMR-CLIP text embeddings. CLIP-image and XLMR-CLIP embeddings are merged and utilized as an input to the latent diffusion process. Also, we conditioned the diffusion process on these embeddings by adding all of them to the time-embedding. Notably, we did not skip the quantization step of the autoencoder during diffusion inference as it leads to an increase in the diversity and the quality of generated images (cf. Figure \ref{fid-clip}).
In total, our model comprises 3.3 B parameters (Table ~\ref{parameters distribution}).

\begin{table}[h]
\small 
  \caption{Kandinsky model parameters.}
  \centering
  \begin{tabular}{lcc}
    \hline
    \textbf{Architecture part} & \textbf{Params} & \textbf{Freeze} \\
    \hline
     Diffusion Mapping & 1B & False \\
     CLIP image encoder (ViT-L14)& 427M & True \\
     CLIP text encoder & 340M & True \\
     Text encoder (XLM-R-L)& 560M & True \\
     Latent Diffusion UNet & 1.22B & False \\
     MoVQ image autoencoder & 67M & True \\
    
    \hline
  \end{tabular}
  \label{parameters distribution}
\end{table}

We observed that the image decoding was our main bottleneck in terms of generated image quality; hence, we developed a Sber-MoVQGAN, our custom implementation of MoVQGAN \cite{zheng2022movq}  with minor modifications. We trained this autoencoder on the LAION HighRes dataset \cite{DBLP:conf/nips/SchuhmannBVGWCC22}, obtaining the SotA results in image reconstruction. We released  the weights and code for these models under an open source licence\footnote{\url{https://github.com/ai-forever/MoVQGAN}}. The comparison of our autoencoder with competitors can be found in Table \ref{movq}.

\begin{table}
\small 
  \caption{Ablation study: FID on COCO-30K validation set on $256 \times 256$ resolution.}
  \label{fid_clip_table}
  \centering
  \begin{tabular}{lcc}
    \hline
    \textbf{Setup} & \textbf{FID-30K} & \textbf{CLIP}\\ 
    \hline
    Diffusion prior with quantization & 9.86 & \textbf{0.287} \\
    Diffusion prior w/o quantization & 9.87 & 0.286\\
    Linear prior & \textbf{8.03} & 0.261\\
    Residual prior & 8.61 & 0.249\\
    No prior & 25.92 & 0.256\\
    
    \hline
  \end{tabular}
\end{table}

\section{Experiments}

\begin{figure}
  \centering
  \includegraphics[width=1.\columnwidth]{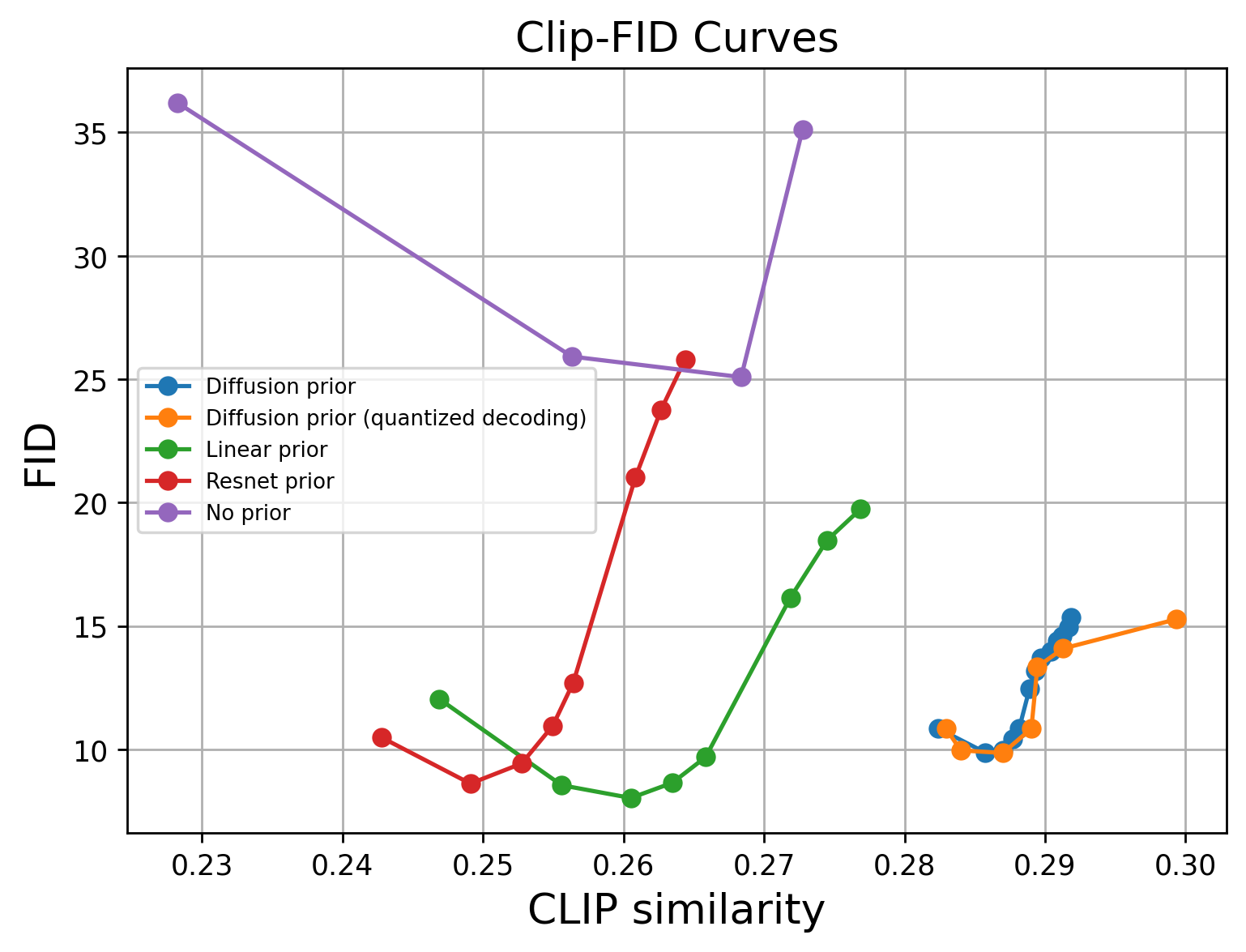}
  \caption{CLIP-FID curves for different setups.}
  \label{fid-clip}
\end{figure}

We sought to evaluate and refine the performance of our proposed latent diffusion architecture in our experimental analysis. To this end, we employed automatic metrics, specifically FID-CLIP curves on the COCO-30K dataset, to obtain the optimal guidance-scale value and compare Kandinsky with competitors (cf. Figure \ref{fid-clip}). Furthermore, we conducted investigations with various image prior setups, exploring the impact of different configurations on the performance. These setups included: no prior, utilizing text embeddings directly; linear prior, implementing one linear layer; ResNet prior, consisting of 18 residual MLP blocks; and transformer diffusion prior.

An essential aspect of our experiments was the exploration of the effect of latent quantization within the MoVQ autoencoder. We examined the outputs with latent quantization, both enabled and disabled, to better comprehend its influence on image generation quality.

To ensure a comprehensive evaluation, we also included an assessment of the IF model \footnote{\label{IF}\url{https://github.com/deep-floyd/IF}}, which is the closest open-source competitor to our proposed model. For this purpose, we computed FID scores for the IF model \footnote{\url{https://github.com/mseitzer/pytorch-fid}} (Table \ref{fid_kandinsky}).

However, we acknowledged the limitations of automatic metrics that become obvious when it comes to capturing user experience nuances. Hence, in addition to the FID-CLIP curves, we conducted a blind human evaluation to obtain insightful feedback and validate the quality of the generated images from the perspective of human perception based on the DrawBench dataset \cite{saharia2022photorealistic}.

The combination of automatic metrics and human evaluation provides a comprehensive assessment of Kandinsky performance, enabling us to make informed decisions about the effectiveness and usability of our proposed image prior to design.
\begin{figure}
  \centering
  \includegraphics[width=1.0\columnwidth]{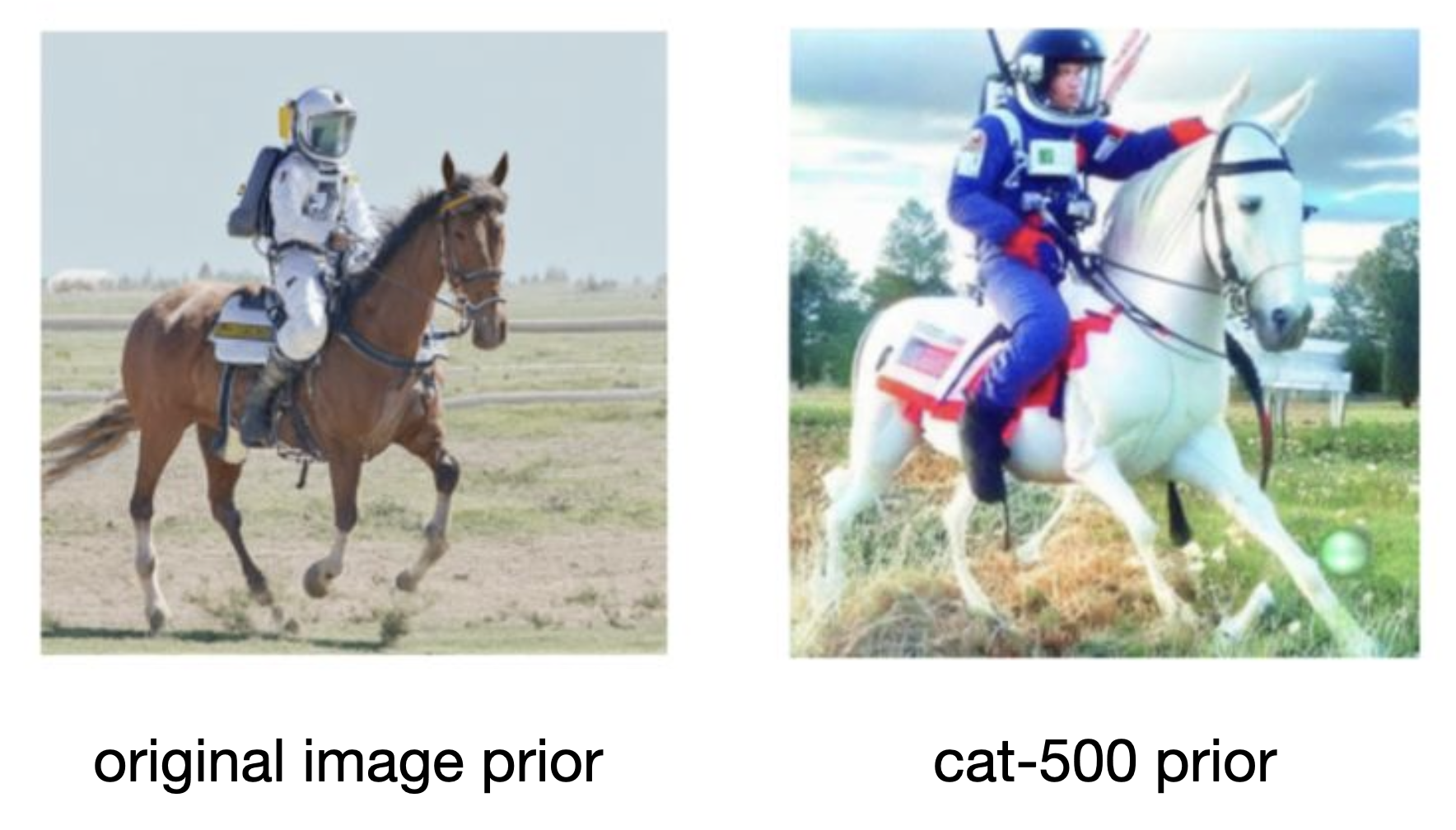}
  \caption{Image generation results with prompt "astronaut riding a horse" for original image prior and linear prior trained on 500 pairs of images with cats.}
  \label{cat prior}
\end{figure}

\begin{table*}[hbt!]
\small
\centering
\caption{Sber-MoVQGAN comparison with competitors on ImageNet dataset.}
\label{tab:model_comparisons}
\begin{tabular}{llllllll}
\hline
\bf Model & \bf  Latent size & \bf  Num Z & \bf  Train steps & \bf  FID $\downarrow$ & \bf  SSIM $\uparrow$ & \bf  PSNR $\uparrow$ & \bf  L1 $\downarrow$ \\
\hline
ViT-VQGAN* & 32x32 & 8192 & 500,000 & 1.28 & - & - & - \\
RQ-VAE* & 8x8x16 & 16384 &  10 epochs & 1.83 & - & - & - \\
Mo-VQGAN* & 16x16x4 & 1024 &  40 epochs & 1.12 & 0.673 & 22.42 & - \\
VQ CompVis & 32x32 & 16384 & 971,043 & 1.34 & 0.650 & 23.85 & 0.0533 \\
KL CompVis & 32x32 & - & 246,803 & 0.968 & 0.692 & 25.11 & 0.0474 \\
Sber-VQGAN & 32x32 & 8192 & 1 epoch & 1.44 & 0.682 & 24.31 & 0.0503 \\
Sber-MoVQGAN 67M & 32x32 & 1024 & 5,000,000 & 1.34 & 0.704 & 25.68 & 0.0451 \\
Sber-MoVQGAN 67M & 32x32 & 16384 & 2,000,000 & 0.965 & 0.725 & 26.45 & 0.0415 \\
Sber-MoVQGAN 102M & 32x32 & 16384 & 2,360,000 & 0.776 & 0.737 & 26.89 & 0.0398 \\
Sber-MoVQGAN 270M & 32x32 & 16384 & 1,330,000 & \textbf{0.686} & \textbf{0.741} & \textbf{27.04} & \textbf{0.0393}\\\hline
\label{movq}
\end{tabular}
\end{table*}

\section{Results}

Our experiments and evaluations have showcased the capabilities of Kandinsky architecture in text-to-image synthesis. Kandinsky achieved the FID score of 8.03 on the COCO-30K validation set at a resolution of 256×256, which puts it in close competition with the state-of-the-art models, and among the top performers within open-source systems. Our methodical ablation studies further dissected the performance of different configurations: quantization of latent codes in MoVQ slightly improves the quality of images (FID 9.86 vs 9.87). The best CLIP score and human-eval score are obtained by diffusion prior. 

The best FID score is achieved using Linear Prior. This configuration stands out with the best FID score of 8.03. It is an intriguing outcome: the simplest linear mapping showcased the best FID, suggesting that there might exist a linear relationship between visual and textual embedding vector spaces. To further scrutinize this hypothesis, we trained a linear mapping on a subset of 500 cat images and termed it the "cat prior". Astonishingly, this mapping displayed high proficiency (cf. Figure~\ref{cat prior}).

\section{Conclusion}

We presented Kandinsky, a system for various image generation and processing tasks based on a novel latent diffusion model. Our model yielded the SotA results among open-sourced systems. Additionally, we provided an extensive ablation study of an image prior to design choices. Our system is equipped with free-to-use interfaces in the form of Web application and Telegram messenger bot. The pre-trained models are available on Hugging Face, and the source code is released under a permissive license enabling various, including commercial, applications of the developed technology. 

In future research, our goal is to investigate the potential of the latest image encoders. We plan to explore the development of more efficient UNet architectures for text-to-image tasks and focus on improving the understanding of textual prompts. Additionally, we aim to experiment with generating images at higher resolutions and to investigate new features extending the model: local image editing by a text prompt, attention reweighting, physics-based generation control, etc. The robustness against generating abusive content remains a crucial concern, warranting the exploration of real-time moderation layers or robust classifiers to mitigate undesirable, e.g. toxic or abusive, outputs.

\section{Limitations} 

The current system produces images that appear natural, however, additional research can be conducted to (1) enhance the semantic coherence between the input text and the generated image, and (2) to improve the absolute values of FID and image quality based on human evaluations.

\section{Ethical Considerations}

We performed multiple efforts to ensure that the generated images do not contain harmful, offensive, or abusive content by (1) cleansing the training dataset from samples that were marked to be harmful/offensive/abusive, and (2) detecting abusive textual prompts. 

While obvious queries, according to our tests, almost never generate abusive content, technically it is not guaranteed that certain carefully engineered prompts may not yield undesirable content. We, therefore, recommend using an additional layer of classifiers, depending on the application, which would filter out the undesired content and/or use image/representation transformation methods tailored to a given application.

\section*{Acknowledgements}
As usual, we would like to thank the anonymous reviewers
for their useful comments. We would also like to
thank Sergey Markov and his team for helpful feedback and discussions, for collaboration in multimodal dataset collecting, labelling and processing.

\bibliographystyle{acl_natbib}
\bibliography{bibliography}

\begin{thebibliography}{41}
\expandafter\ifx\csname natexlab\endcsname\relax\def\natexlab#1{#1}\fi

\bibitem[{Balaji et~al.(2022)Balaji, Nah, Huang, Vahdat, Song, Kreis, Aittala,
  Aila, Laine, Catanzaro, Karras, and Liu}]{balaji2023ediffi}
Yogesh Balaji, Seungjun Nah, Xun Huang, Arash Vahdat, Jiaming Song, Karsten
  Kreis, Miika Aittala, Timo Aila, Samuli Laine, Bryan Catanzaro, Tero Karras,
  and Ming{-}Yu Liu. 2022.
\newblock \href {https://doi.org/10.48550/arXiv.2211.01324} {ediff-i:
  Text-to-image diffusion models with an ensemble of expert denoisers}.

\bibitem[{Batzolis et~al.(2021)Batzolis, Stanczuk, Sch{\"{o}}nlieb, and
  Etmann}]{batzolis2021conditional}
Georgios Batzolis, Jan Stanczuk, Carola{-}Bibiane Sch{\"{o}}nlieb, and
  Christian Etmann. 2021.
\newblock \href {http://arxiv.org/abs/2111.13606} {Conditional image generation
  with score-based diffusion models}.
\newblock \emph{CoRR}, abs/2111.13606.

\bibitem[{Blattmann et~al.(2023)Blattmann, Rombach, Ling, Dockhorn, Kim,
  Fidler, and Kreis}]{DBLP:journals/corr/abs-2304-08818}
Andreas Blattmann, Robin Rombach, Huan Ling, Tim Dockhorn, Seung~Wook Kim,
  Sanja Fidler, and Karsten Kreis. 2023.
\newblock \href {https://doi.org/10.48550/arXiv.2304.08818} {Align your
  latents: High-resolution video synthesis with latent diffusion models}.
\newblock \emph{CoRR}, abs/2304.08818.

\bibitem[{Chen et~al.(2023)Chen, Chen, Jiao, and
  Jia}]{DBLP:journals/corr/abs-2303-13873}
Rui Chen, Yongwei Chen, Ningxin Jiao, and Kui Jia. 2023.
\newblock \href {https://doi.org/10.48550/arXiv.2303.13873} {Fantasia3d:
  Disentangling geometry and appearance for high-quality text-to-3d content
  creation}.
\newblock \emph{CoRR}, abs/2303.13873.

\bibitem[{Conneau et~al.(2020)Conneau, Khandelwal, Goyal, Chaudhary, Wenzek,
  Guzm{\'{a}}n, Grave, Ott, Zettlemoyer, and
  Stoyanov}]{DBLP:conf/acl/ConneauKGCWGGOZ20}
Alexis Conneau, Kartikay Khandelwal, Naman Goyal, Vishrav Chaudhary, Guillaume
  Wenzek, Francisco Guzm{\'{a}}n, Edouard Grave, Myle Ott, Luke Zettlemoyer,
  and Veselin Stoyanov. 2020.
\newblock \href {https://doi.org/10.18653/v1/2020.acl-main.747} {Unsupervised
  cross-lingual representation learning at scale}.
\newblock In \emph{Proceedings of the 58th Annual Meeting of the Association
  for Computational Linguistics, {ACL} 2020, Online, July 5-10, 2020}, pages
  8440--8451. Association for Computational Linguistics.

\bibitem[{Dhariwal and Nichol(2021)}]{dhariwal2021diffusion}
Prafulla Dhariwal and Alexander~Quinn Nichol. 2021.
\newblock \href
  {https://proceedings.neurips.cc/paper/2021/hash/49ad23d1ec9fa4bd8d77d02681df5cfa-Abstract.html}
  {Diffusion models beat gans on image synthesis}.
\newblock In \emph{Advances in Neural Information Processing Systems 34: Annual
  Conference on Neural Information Processing Systems 2021, NeurIPS 2021,
  December 6-14, 2021, virtual}, pages 8780--8794.

\bibitem[{Ding et~al.(2021)Ding, Yang, Hong, Zheng, Zhou, Yin, Lin, Zou, Shao,
  Yang, and Tang}]{DBLP:conf/nips/DingYHZZYLZSYT21}
Ming Ding, Zhuoyi Yang, Wenyi Hong, Wendi Zheng, Chang Zhou, Da~Yin, Junyang
  Lin, Xu~Zou, Zhou Shao, Hongxia Yang, and Jie Tang. 2021.
\newblock \href
  {https://proceedings.neurips.cc/paper/2021/hash/a4d92e2cd541fca87e4620aba658316d-Abstract.html}
  {Cogview: Mastering text-to-image generation via transformers}.
\newblock In \emph{Advances in Neural Information Processing Systems 34: Annual
  Conference on Neural Information Processing Systems 2021, NeurIPS 2021,
  December 6-14, 2021, virtual}, pages 19822--19835.

\bibitem[{Esser et~al.(2023)Esser, Chiu, Atighehchian, Granskog, and
  Germanidis}]{DBLP:journals/corr/abs-2302-03011}
Patrick Esser, Johnathan Chiu, Parmida Atighehchian, Jonathan Granskog, and
  Anastasis Germanidis. 2023.
\newblock \href {https://doi.org/10.48550/arXiv.2302.03011} {Structure and
  content-guided video synthesis with diffusion models}.
\newblock \emph{CoRR}, abs/2302.03011.

\bibitem[{Goodfellow et~al.(2014)Goodfellow, Pouget{-}Abadie, Mirza, Xu,
  Warde{-}Farley, Ozair, Courville, and Bengio}]{NIPS2014_5ca3e9b1}
Ian~J. Goodfellow, Jean Pouget{-}Abadie, Mehdi Mirza, Bing Xu, David
  Warde{-}Farley, Sherjil Ozair, Aaron~C. Courville, and Yoshua Bengio. 2014.
\newblock \href
  {https://proceedings.neurips.cc/paper/2014/hash/5ca3e9b122f61f8f06494c97b1afccf3-Abstract.html}
  {Generative adversarial nets}.
\newblock In \emph{Advances in Neural Information Processing Systems 27: Annual
  Conference on Neural Information Processing Systems 2014, December 8-13 2014,
  Montreal, Quebec, Canada}, pages 2672--2680.

\bibitem[{Hertz et~al.(2023)Hertz, Mokady, Tenenbaum, Aberman, Pritch, and
  Cohen{-}Or}]{DBLP:conf/iclr/HertzMTAPC23}
Amir Hertz, Ron Mokady, Jay Tenenbaum, Kfir Aberman, Yael Pritch, and Daniel
  Cohen{-}Or. 2023.
\newblock \href {https://openreview.net/pdf?id=\_CDixzkzeyb} {Prompt-to-prompt
  image editing with cross-attention control}.
\newblock In \emph{The Eleventh International Conference on Learning
  Representations, {ICLR} 2023, Kigali, Rwanda, May 1-5, 2023}. OpenReview.net.

\bibitem[{Ho et~al.(2022{\natexlab{a}})Ho, Chan, Saharia, Whang, Gao,
  Gritsenko, Kingma, Poole, Norouzi, Fleet, and
  Salimans}]{DBLP:journals/corr/abs-2210-02303}
Jonathan Ho, William Chan, Chitwan Saharia, Jay Whang, Ruiqi Gao, Alexey~A.
  Gritsenko, Diederik~P. Kingma, Ben Poole, Mohammad Norouzi, David~J. Fleet,
  and Tim Salimans. 2022{\natexlab{a}}.
\newblock \href {https://doi.org/10.48550/arXiv.2210.02303} {Imagen video: High
  definition video generation with diffusion models}.
\newblock \emph{CoRR}, abs/2210.02303.

\bibitem[{Ho et~al.(2020)Ho, Jain, and Abbeel}]{ho2020denoising}
Jonathan Ho, Ajay Jain, and Pieter Abbeel. 2020.
\newblock \href
  {https://proceedings.neurips.cc/paper/2020/hash/4c5bcfec8584af0d967f1ab10179ca4b-Abstract.html}
  {Denoising diffusion probabilistic models}.
\newblock In \emph{Advances in Neural Information Processing Systems 33: Annual
  Conference on Neural Information Processing Systems 2020, NeurIPS 2020,
  December 6-12, 2020, virtual}.

\bibitem[{Ho and Salimans(2022)}]{ho2021classifierfree}
Jonathan Ho and Tim Salimans. 2022.
\newblock \href {https://doi.org/10.48550/arXiv.2207.12598} {Classifier-free
  diffusion guidance}.
\newblock volume abs/2207.12598.

\bibitem[{Ho et~al.(2022{\natexlab{b}})Ho, Salimans, Gritsenko, Chan, Norouzi,
  and Fleet}]{DBLP:conf/nips/HoSGC0F22}
Jonathan Ho, Tim Salimans, Alexey~A. Gritsenko, William Chan, Mohammad Norouzi,
  and David~J. Fleet. 2022{\natexlab{b}}.
\newblock \href
  {http://papers.nips.cc/paper\_files/paper/2022/hash/39235c56aef13fb05a6adc95eb9d8d66-Abstract-Conference.html}
  {Video diffusion models}.
\newblock In \emph{NeurIPS}.

\bibitem[{Kang et~al.(2023)Kang, Zhu, Zhang, Park, Shechtman, Paris, and
  Park}]{kang2023scaling}
Minguk Kang, Jun{-}Yan Zhu, Richard Zhang, Jaesik Park, Eli Shechtman, Sylvain
  Paris, and Taesung Park. 2023.
\newblock \href {https://doi.org/10.1109/CVPR52729.2023.00976} {Scaling up gans
  for text-to-image synthesis}.
\newblock In \emph{{IEEE/CVF} Conference on Computer Vision and Pattern
  Recognition, {CVPR} 2023, Vancouver, BC, Canada, June 17-24, 2023}, pages
  10124--10134. {IEEE}.

\bibitem[{Karras et~al.(2023)Karras, Holynski, Wang, and
  Kemelmacher{-}Shlizerman}]{karras2023dreampose}
Johanna Karras, Aleksander Holynski, Ting{-}Chun Wang, and Ira
  Kemelmacher{-}Shlizerman. 2023.
\newblock \href {https://doi.org/10.48550/arXiv.2304.06025} {Dreampose: Fashion
  image-to-video synthesis via stable diffusion}.

\bibitem[{Li et~al.(2023)Li, Liu, Wu, Mu, Yang, Gao, Li, and
  Lee}]{DBLP:journals/corr/abs-2301-07093}
Yuheng Li, Haotian Liu, Qingyang Wu, Fangzhou Mu, Jianwei Yang, Jianfeng Gao,
  Chunyuan Li, and Yong~Jae Lee. 2023.
\newblock \href {https://doi.org/10.48550/arXiv.2301.07093} {{GLIGEN:} open-set
  grounded text-to-image generation}.
\newblock \emph{CoRR}, abs/2301.07093.

\bibitem[{Liew et~al.(2022)Liew, Yan, Zhou, and
  Feng}]{DBLP:journals/corr/abs-2210-16056}
Jun~Hao Liew, Hanshu Yan, Daquan Zhou, and Jiashi Feng. 2022.
\newblock \href {https://doi.org/10.48550/arXiv.2210.16056} {Magicmix: Semantic
  mixing with diffusion models}.
\newblock \emph{CoRR}, abs/2210.16056.

\bibitem[{Lin et~al.(2022)Lin, Gao, Tang, Takikawa, Zeng, Huang, Kreis, Fidler,
  Liu, and Lin}]{DBLP:journals/corr/abs-2211-10440}
Chen{-}Hsuan Lin, Jun Gao, Luming Tang, Towaki Takikawa, Xiaohui Zeng, Xun
  Huang, Karsten Kreis, Sanja Fidler, Ming{-}Yu Liu, and Tsung{-}Yi Lin. 2022.
\newblock \href {https://doi.org/10.48550/arXiv.2211.10440} {Magic3d:
  High-resolution text-to-3d content creation}.
\newblock \emph{CoRR}, abs/2211.10440.

\bibitem[{Lu et~al.(2023)Lu, Liu, and Kong}]{DBLP:journals/corr/abs-2307-12493}
Shilin Lu, Yanzhu Liu, and Adams~Wai{-}Kin Kong. 2023.
\newblock \href {https://doi.org/10.48550/arXiv.2307.12493} {{TF-ICON:}
  diffusion-based training-free cross-domain image composition}.
\newblock \emph{CoRR}, abs/2307.12493.

\bibitem[{Luo et~al.(2023)Luo, Chen, Zhang, Huang, Wang, Shen, Zhao, Zhou, and
  Tan}]{DBLP:journals/corr/abs-2303-08320}
Zhengxiong Luo, Dayou Chen, Yingya Zhang, Yan Huang, Liang Wang, Yujun Shen,
  Deli Zhao, Jingren Zhou, and Tieniu Tan. 2023.
\newblock \href {https://doi.org/10.48550/arXiv.2303.08320} {Videofusion:
  Decomposed diffusion models for high-quality video generation}.
\newblock \emph{CoRR}, abs/2303.08320.

\bibitem[{Mou et~al.(2023)Mou, Wang, Song, Shan, and
  Zhang}]{DBLP:journals/corr/abs-2307-02421}
Chong Mou, Xintao Wang, Jiechong Song, Ying Shan, and Jian Zhang. 2023.
\newblock \href {https://doi.org/10.48550/arXiv.2307.02421} {Dragondiffusion:
  Enabling drag-style manipulation on diffusion models}.
\newblock \emph{CoRR}, abs/2307.02421.

\bibitem[{Nichol and Dhariwal(2021)}]{nichol2021improved}
Alexander~Quinn Nichol and Prafulla Dhariwal. 2021.
\newblock \href {http://proceedings.mlr.press/v139/nichol21a.html} {Improved
  denoising diffusion probabilistic models}.
\newblock In \emph{Proceedings of the 38th International Conference on Machine
  Learning, {ICML} 2021, 18-24 July 2021, Virtual Event}, volume 139 of
  \emph{Proceedings of Machine Learning Research}, pages 8162--8171. {PMLR}.

\bibitem[{Nichol et~al.(2022)Nichol, Dhariwal, Ramesh, Shyam, Mishkin, McGrew,
  Sutskever, and Chen}]{DBLP:conf/icml/NicholDRSMMSC22}
Alexander~Quinn Nichol, Prafulla Dhariwal, Aditya Ramesh, Pranav Shyam, Pamela
  Mishkin, Bob McGrew, Ilya Sutskever, and Mark Chen. 2022.
\newblock \href {https://proceedings.mlr.press/v162/nichol22a.html} {{GLIDE:}
  towards photorealistic image generation and editing with text-guided
  diffusion models}.
\newblock In \emph{International Conference on Machine Learning, {ICML} 2022,
  17-23 July 2022, Baltimore, Maryland, {USA}}, volume 162 of \emph{Proceedings
  of Machine Learning Research}, pages 16784--16804. {PMLR}.

\bibitem[{Parmar et~al.(2023)Parmar, Singh, Zhang, Li, Lu, and
  Zhu}]{DBLP:conf/siggraph/ParmarS0LLZ23}
Gaurav Parmar, Krishna~Kumar Singh, Richard Zhang, Yijun Li, Jingwan Lu, and
  Jun{-}Yan Zhu. 2023.
\newblock \href {https://doi.org/10.1145/3588432.3591513} {Zero-shot
  image-to-image translation}.
\newblock In \emph{{ACM} {SIGGRAPH} 2023 Conference Proceedings, {SIGGRAPH}
  2023, Los Angeles, CA, USA, August 6-10, 2023}, pages 11:1--11:11. {ACM}.

\bibitem[{Peebles and Xie(2022)}]{peebles2022scalable}
William Peebles and Saining Xie. 2022.
\newblock \href {https://doi.org/10.48550/arXiv.2212.09748} {Scalable diffusion
  models with transformers}.
\newblock \emph{CoRR}, abs/2212.09748.

\bibitem[{Poole et~al.(2023)Poole, Jain, Barron, and
  Mildenhall}]{DBLP:conf/iclr/PooleJBM23}
Ben Poole, Ajay Jain, Jonathan~T. Barron, and Ben Mildenhall. 2023.
\newblock \href {https://openreview.net/pdf?id=FjNys5c7VyY} {Dreamfusion:
  Text-to-3d using 2d diffusion}.
\newblock In \emph{The Eleventh International Conference on Learning
  Representations, {ICLR} 2023, Kigali, Rwanda, May 1-5, 2023}. OpenReview.net.

\bibitem[{Radford et~al.(2021)Radford, Kim, Hallacy, Ramesh, Goh, Agarwal,
  Sastry, Askell, Mishkin, Clark, Krueger, and
  Sutskever}]{DBLP:conf/icml/RadfordKHRGASAM21}
Alec Radford, Jong~Wook Kim, Chris Hallacy, Aditya Ramesh, Gabriel Goh,
  Sandhini Agarwal, Girish Sastry, Amanda Askell, Pamela Mishkin, Jack Clark,
  Gretchen Krueger, and Ilya Sutskever. 2021.
\newblock \href {http://proceedings.mlr.press/v139/radford21a.html} {Learning
  transferable visual models from natural language supervision}.
\newblock In \emph{Proceedings of the 38th International Conference on Machine
  Learning, {ICML} 2021, 18-24 July 2021, Virtual Event}, volume 139 of
  \emph{Proceedings of Machine Learning Research}, pages 8748--8763. {PMLR}.

\bibitem[{Raffel et~al.(2020)Raffel, Shazeer, Roberts, Lee, Narang, Matena,
  Zhou, Li, and Liu}]{raffel2020exploring}
Colin Raffel, Noam Shazeer, Adam Roberts, Katherine Lee, Sharan Narang, Michael
  Matena, Yanqi Zhou, Wei Li, and Peter~J. Liu. 2020.
\newblock \href {http://jmlr.org/papers/v21/20-074.html} {Exploring the limits
  of transfer learning with a unified text-to-text transformer}.
\newblock \emph{J. Mach. Learn. Res.}, 21:140:1--140:67.

\bibitem[{Ramesh et~al.(2022)Ramesh, Dhariwal, Nichol, Chu, and
  Chen}]{ramesh2022hierarchical}
Aditya Ramesh, Prafulla Dhariwal, Alex Nichol, Casey Chu, and Mark Chen. 2022.
\newblock \href {https://doi.org/10.48550/arXiv.2204.06125} {Hierarchical
  text-conditional image generation with {CLIP} latents}.

\bibitem[{Ramesh et~al.(2021)Ramesh, Pavlov, Goh, Gray, Voss, Radford, Chen,
  and Sutskever}]{DBLP:conf/icml/RameshPGGVRCS21}
Aditya Ramesh, Mikhail Pavlov, Gabriel Goh, Scott Gray, Chelsea Voss, Alec
  Radford, Mark Chen, and Ilya Sutskever. 2021.
\newblock \href {http://proceedings.mlr.press/v139/ramesh21a.html} {Zero-shot
  text-to-image generation}.
\newblock In \emph{Proceedings of the 38th International Conference on Machine
  Learning, {ICML} 2021, 18-24 July 2021, Virtual Event}, volume 139 of
  \emph{Proceedings of Machine Learning Research}, pages 8821--8831. {PMLR}.

\bibitem[{Rombach et~al.(2022)Rombach, Blattmann, Lorenz, Esser, and
  Ommer}]{rombach2022highresolution}
Robin Rombach, Andreas Blattmann, Dominik Lorenz, Patrick Esser, and
  Bj{\"{o}}rn Ommer. 2022.
\newblock \href {https://doi.org/10.1109/CVPR52688.2022.01042} {High-resolution
  image synthesis with latent diffusion models}.
\newblock In \emph{{IEEE/CVF} Conference on Computer Vision and Pattern
  Recognition, {CVPR} 2022, New Orleans, LA, USA, June 18-24, 2022}, pages
  10674--10685. {IEEE}.

\bibitem[{Ruiz et~al.(2023)Ruiz, Li, Jampani, Pritch, Rubinstein, and
  Aberman}]{ruiz2022dreambooth}
Nataniel Ruiz, Yuanzhen Li, Varun Jampani, Yael Pritch, Michael Rubinstein, and
  Kfir Aberman. 2023.
\newblock \href {https://doi.org/10.1109/CVPR52729.2023.02155} {Dreambooth:
  Fine tuning text-to-image diffusion models for subject-driven generation}.
\newblock In \emph{{IEEE/CVF} Conference on Computer Vision and Pattern
  Recognition, {CVPR} 2023, Vancouver, BC, Canada, June 17-24, 2023}, pages
  22500--22510. {IEEE}.

\bibitem[{Saharia et~al.(2022{\natexlab{a}})Saharia, Chan, Chang, Lee, Ho,
  Salimans, Fleet, and Norouzi}]{saharia2022palette}
Chitwan Saharia, William Chan, Huiwen Chang, Chris~A. Lee, Jonathan Ho, Tim
  Salimans, David~J. Fleet, and Mohammad Norouzi. 2022{\natexlab{a}}.
\newblock \href {https://doi.org/10.1145/3528233.3530757} {Palette:
  Image-to-image diffusion models}.
\newblock In \emph{{SIGGRAPH} '22: Special Interest Group on Computer Graphics
  and Interactive Techniques Conference, Vancouver, BC, Canada, August 7 - 11,
  2022}, pages 15:1--15:10. {ACM}.

\bibitem[{Saharia et~al.(2022{\natexlab{b}})Saharia, Chan, Saxena, Li, Whang,
  Denton, Ghasemipour, Lopes, Ayan, Salimans, Ho, Fleet, and
  Norouzi}]{saharia2022photorealistic}
Chitwan Saharia, William Chan, Saurabh Saxena, Lala Li, Jay Whang, Emily~L.
  Denton, Seyed Kamyar~Seyed Ghasemipour, Raphael~Gontijo Lopes, Burcu~Karagol
  Ayan, Tim Salimans, Jonathan Ho, David~J. Fleet, and Mohammad Norouzi.
  2022{\natexlab{b}}.
\newblock \href
  {http://papers.nips.cc/paper\_files/paper/2022/hash/ec795aeadae0b7d230fa35cbaf04c041-Abstract-Conference.html}
  {Photorealistic text-to-image diffusion models with deep language
  understanding}.

\bibitem[{Schuhmann et~al.(2022)Schuhmann, Beaumont, Vencu, Gordon, Wightman,
  Cherti, Coombes, Katta, Mullis, Wortsman, Schramowski, Kundurthy, Crowson,
  Schmidt, Kaczmarczyk, and Jitsev}]{DBLP:conf/nips/SchuhmannBVGWCC22}
Christoph Schuhmann, Romain Beaumont, Richard Vencu, Cade Gordon, Ross
  Wightman, Mehdi Cherti, Theo Coombes, Aarush Katta, Clayton Mullis, Mitchell
  Wortsman, Patrick Schramowski, Srivatsa Kundurthy, Katherine Crowson, Ludwig
  Schmidt, Robert Kaczmarczyk, and Jenia Jitsev. 2022.
\newblock \href
  {http://papers.nips.cc/paper\_files/paper/2022/hash/a1859debfb3b59d094f3504d5ebb6c25-Abstract-Datasets\_and\_Benchmarks.html}
  {{LAION-5B:} an open large-scale dataset for training next generation
  image-text models}.
\newblock In \emph{NeurIPS}.

\bibitem[{Singer et~al.(2023)Singer, Polyak, Hayes, Yin, An, Zhang, Hu, Yang,
  Ashual, Gafni, Parikh, Gupta, and Taigman}]{DBLP:conf/iclr/SingerPH00ZHYAG23}
Uriel Singer, Adam Polyak, Thomas Hayes, Xi~Yin, Jie An, Songyang Zhang, Qiyuan
  Hu, Harry Yang, Oron Ashual, Oran Gafni, Devi Parikh, Sonal Gupta, and Yaniv
  Taigman. 2023.
\newblock \href {https://openreview.net/pdf?id=nJfylDvgzlq} {Make-a-video:
  Text-to-video generation without text-video data}.
\newblock In \emph{The Eleventh International Conference on Learning
  Representations, {ICLR} 2023, Kigali, Rwanda, May 1-5, 2023}. OpenReview.net.

\bibitem[{Tang et~al.(2023)Tang, Wang, Zhang, Zhang, Yi, Ma, and
  Chen}]{DBLP:journals/corr/abs-2303-14184}
Junshu Tang, Tengfei Wang, Bo~Zhang, Ting Zhang, Ran Yi, Lizhuang Ma, and Dong
  Chen. 2023.
\newblock \href {https://doi.org/10.48550/arXiv.2303.14184} {Make-it-3d:
  High-fidelity 3d creation from {A} single image with diffusion prior}.
\newblock \emph{CoRR}, abs/2303.14184.

\bibitem[{Xue et~al.(2021)Xue, Constant, Roberts, Kale, Al{-}Rfou, Siddhant,
  Barua, and Raffel}]{DBLP:conf/naacl/XueCRKASBR21}
Linting Xue, Noah Constant, Adam Roberts, Mihir Kale, Rami Al{-}Rfou, Aditya
  Siddhant, Aditya Barua, and Colin Raffel. 2021.
\newblock \href {https://doi.org/10.18653/v1/2021.naacl-main.41} {mt5: {A}
  massively multilingual pre-trained text-to-text transformer}.
\newblock In \emph{Proceedings of the 2021 Conference of the North American
  Chapter of the Association for Computational Linguistics: Human Language
  Technologies, {NAACL-HLT} 2021, Online, June 6-11, 2021}, pages 483--498.
  Association for Computational Linguistics.

\bibitem[{Yu et~al.(2022)Yu, Xu, Koh, Luong, Baid, Wang, Vasudevan, Ku, Yang,
  Ayan, Hutchinson, Han, Parekh, Li, Zhang, Baldridge, and
  Wu}]{DBLP:journals/tmlr/YuXKLBWVKYAHHPLZBW22}
Jiahui Yu, Yuanzhong Xu, Jing~Yu Koh, Thang Luong, Gunjan Baid, Zirui Wang,
  Vijay Vasudevan, Alexander Ku, Yinfei Yang, Burcu~Karagol Ayan, Ben
  Hutchinson, Wei Han, Zarana Parekh, Xin Li, Han Zhang, Jason Baldridge, and
  Yonghui Wu. 2022.
\newblock \href {https://openreview.net/forum?id=AFDcYJKhND} {Scaling
  autoregressive models for content-rich text-to-image generation}.
\newblock \emph{Trans. Mach. Learn. Res.}, 2022.

\bibitem[{Zheng et~al.(2022)Zheng, Vuong, Cai, and Phung}]{zheng2022movq}
Chuanxia Zheng, Tung{-}Long Vuong, Jianfei Cai, and Dinh Phung. 2022.
\newblock \href
  {http://papers.nips.cc/paper\_files/paper/2022/hash/94840c41497ead6a84f493f029eba7fa-Abstract-Conference.html}
  {Movq: Modulating quantized vectors for high-fidelity image generation}.
\newblock In \emph{NeurIPS}.

\end{thebibliography}

\clearpage
\appendix

\section{Additional generation examples}

\begin{figure}[htp!]
  \centering
  \includegraphics[width=1.95\columnwidth]{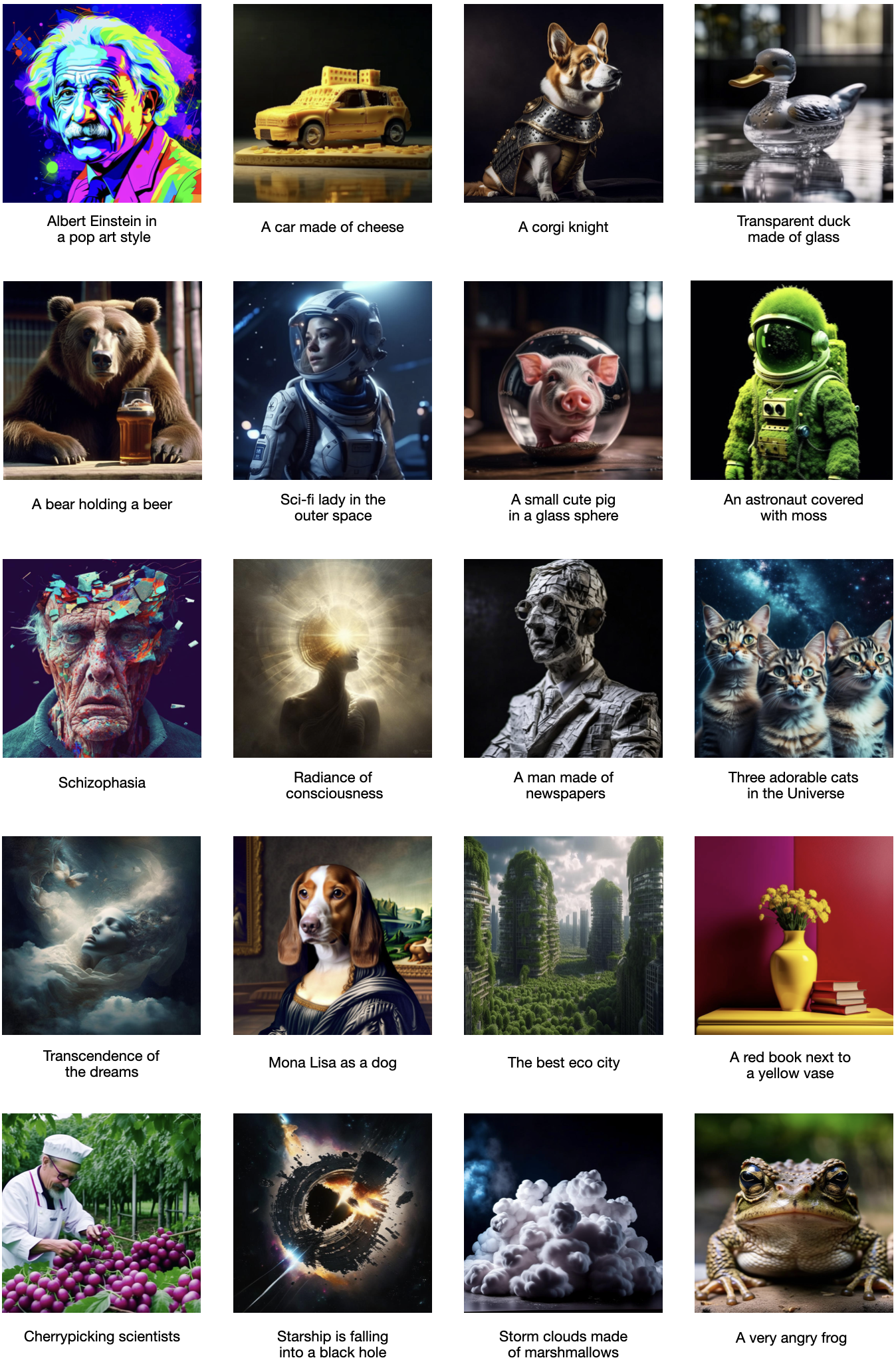}
  \label{examples}
\end{figure}

\end{document}